\documentclass{article}

\usepackage[preprint]{nips_2018}

\usepackage[utf8]{inputenc} % allow utf-8 input
\usepackage[T1]{fontenc}    % use 8-bit T1 fonts
\usepackage{hyperref}       % hyperlinks
\usepackage{url}            % simple URL typesetting
\usepackage{booktabs}       % professional-quality tables
\usepackage{amsfonts}       % blackboard math symbols
\usepackage{nicefrac}       % compact symbols for 1/2, etc.
\usepackage{microtype}      % microtypography
\usepackage{graphicx}
\usepackage{amsmath}
\usepackage{amsthm}
\usepackage{mathtools}
\usepackage{algorithm}
\usepackage{algpseudocode}
\usepackage[title]{appendix}
\newcommand{\ones}{\mathbf 1}
\newcommand{\reals}{{\mbox{\bf R}}}

\newcommand{\Expect}{\mathbb{E}}
\newcommand{\Prob}{\mathbb{P}}
\newcommand{\eg}{{\it e.g.}}
\newcommand{\ie}{{\it i.e.}}

\theoremstyle{definition}
\newtheorem{definition}{Definition}[section]

\theoremstyle{definition}

\theoremstyle{definition}
\newtheorem{theorem}{Theorem}[section]

\theoremstyle{definition}
\newtheorem{exmp}{Example}[section]

\title{Improved Training with Curriculum GANs}
\author{
  Rishi Sharma\thanks{Correspondence to \texttt{rsh@stanford.edu}.} \;\; Shane Barratt  \;\; Stefano Ermon \;\; Vijay Pande\\
  Stanford University\\
}

\begin{document}

\maketitle

\begin{abstract}
In this paper we introduce Curriculum GANs, a curriculum learning strategy for training Generative Adversarial Networks that increases the strength of the discriminator over the course of training, thereby making the learning task progressively more difficult for the generator. We demonstrate that this strategy is key to obtaining state-of-the-art results in image generation. We also show evidence that this strategy may be broadly applicable to improving GAN training in other data modalities.
\end{abstract}

\section{Introduction}\label{introduction}
Generative Adversarial Networks (GANs) are an innovative approach to generative modeling that cast the problem of producing synthetic data as a game between two adversaries: a generator, which seeks to produce samples from the same distribution as the data, and a discriminator, whose job is to distinguish between real and generated data \citep{goodfellow2014gans}. In practice, by implementing the generator and discriminator as competing deep neural networks, each trained via stochastic gradient methods, GANs are capable of producing plausible synthetic data across a wide diversity of data modalities, including natural images \citep{radford2015dcgan,karras2017progressive}, natural language \citep{yu16seqgan,press17language-gen,fedus18maskgan}, medical records \citep{esteban2017medical} and molecules \citep{sanchez-lengeling2017organic}.

\begin{figure}
\centering
\includegraphics[width=\columnwidth]{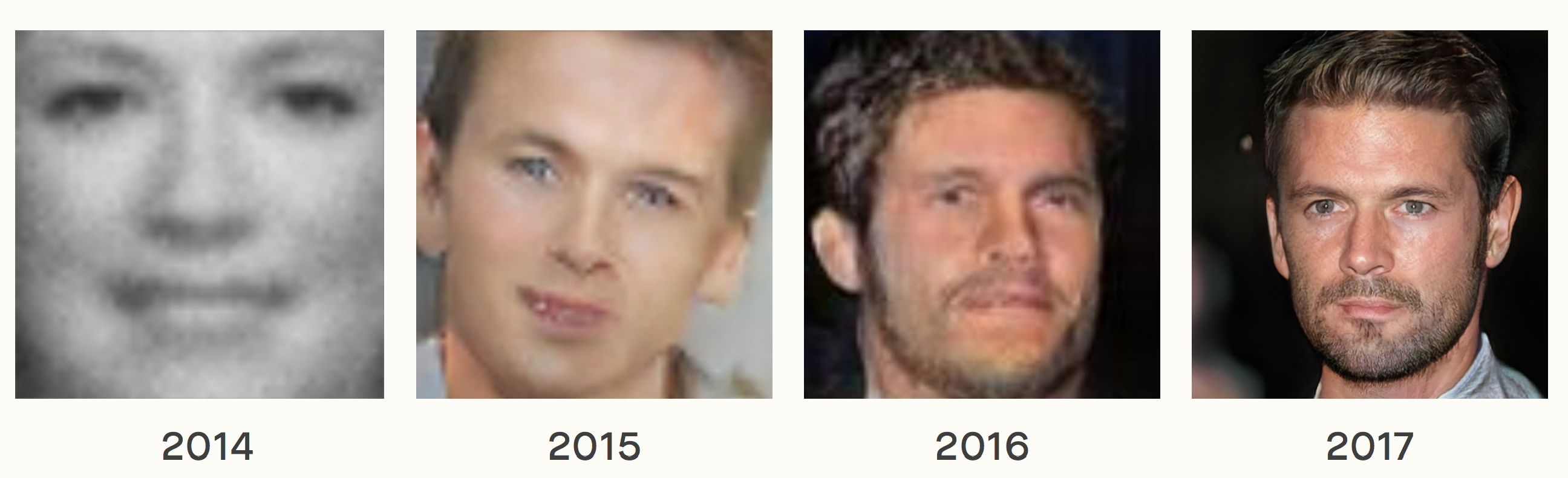}
\caption{Progress in GANs from 2014 to 2017 (image source: \citet{brundage18maliciousAI})}
\label{gan-progress}
\end{figure}

Despite these successes, training GANs via stochastic gradient methods remains unstable and prone to a variety of failure modes. This has led to a proliferation of work that focuses on improving the quality of the output of GANs by stabilizing the training procedure \citep{salimans2016improved,poole16improved, warde2016improving,gulrajani2017improved,karras2017progressive}. Through incremental successes, the deep generative modeling community has amassed a set of tips, tricks and hacks that have made training GANs easier\footnote{https://github.com/soumith/ganhacks}$^,$\footnote{https://medium.com/@utk.is.here/keep-calm-and-train-a-gan-pitfalls-and-tips-on-training-generative-adversarial-networks-edd529764aa9}. A patchwork of methods has emerged; specialized techniques---uniquely suited to different domains---have enabled GANs both to expand to a diversity of new domains and to continuously improve on the state-of-the-art in classical areas such as image generation. These efforts have been successful, as evidenced by the improved quality of GAN-generated faces in recent years shown in Figure~\ref{gan-progress}.

Nevertheless, a lack of strong evaluative metrics for GANs {\citep{theis2015note,barratt18noteIS}} has made it difficult to isolate precisely which methods have produced improvements or to reliably predict the regimes in which those improvements will hold. In this paper, we clarify which elements in the GAN ``bag of tricks'' have improved the application of GANs to image generation. In particular, we focus on the work of~\citet{karras2017progressive}, which gained widespread attention for its high-quality rendering of fake celebrity images via the layer-wise growing of a deep convolutional network. We simplify their model by introducing the concept of a {\it GAN curriculum}, and we argue that a well-crafted curriculum, one that gradually increases the capabilities of the discriminator, is the key to obtaining their state-of-the-art results. We remove the need for the complicated layer-wise training in their model, dramatically reducing the complexity of their setup, and still obtain results of the same (high) quality. As such, we argue that their instantiation of our framework was the primary contributing factor to their high quality images.

We also obtain preliminary results that indicate this technique may be generally applicable in broader GAN settings and capable of improving GANs beyond image generation. In Section~\ref{discussion-related}, we point to literature in natural language generation and text-to-image synthesis that takes advantage of related techniques, suggesting that informal variants of the method we formalize have already led to success in training GANs in other areas.

A recent paper by~\citet{lucic2017gans} rebuked many of the claims of improved GAN training and performance by conducting a large-scale, multi-faceted study and finding little evidence that newer training setups outperform the original GAN of~\citet{goodfellow2014gans}. The paper accurately argued that improving GANs generally is an altogether more difficult task than improving GANs designed for a specific purpose, \eg, image generation. Despite the advances made by GANs using the "bag of tricks", we have failed the challenge of building training procedures that outperform the original setup broadly across many data modalities and tasks. Understanding why the bag of tricks has worked in specific environments is essential to eventually succeeding in creating training methods that in fact improve GANs in the general setting. As such, in addition to their utility in image generation, we believe that our findings represent an important step towards improving GANs generally.

\section{Designing a Curriculum for Generative Adversarial Networks}
\label{method}
In this section, we describe a general method for GAN training that helps to prevent instabilities during training and thus improve the quality of the final learned generator parameters.
The main idea behind our method is to construct a training regimen for the generator that consists of increasingly difficult tasks.
This allows the sophistication of the generator to gradually increase throughout training, rather than aiming for full sophistication at the outset.
This method is similar to that of a \emph{curriculum} in supervised learning, where one orders the training examples presented to a learning algorithm according to some measure of difficulty~\citep{bengio2009curriculum}.
Despite the conceptual similarity, the methods are in fact quite different.
Under our approach, it is not the difficulty of the training examples presented to either network, but rather the capacity, and hence strength, of the discriminator that is increased throughout training.

\subsection{Preliminaries}
The goal in generative modeling is to learn the probability distribution $\Prob_\text{r}$ of some random variable $x\in\mathcal{X}$ from a dataset of samples $x_1,\ldots,x_N$ drawn from $\Prob_\text{r}$.
In GANs (and other deep generative models), it is common practice to define a random variable $z\in\mathcal{Z}$ with a known, fixed distribution $p(z)$.
The generator is then defined as a parametric function $g_\theta:\mathcal{Z}\rightarrow\mathcal{X}$ that transforms $z$ into artificial samples.
Implicitly, $g_\theta(z)$ is now a random variable with a distribution that we denote $\Prob_\text{g}$.
It is easy to sample from $\Prob_\text{g}$, as all we need to do is sample $z\sim p(z)$ and then emit $g_\theta(z)$.
The goal is to learn the parameters $\theta$ of the generator so that $\Prob_\text{g}$ is as similar as possible to $\Prob_\text{r}$. 
To do this, we first define a discriminator $f_w:\mathcal{X}\rightarrow\reals$ for $w\in\mathcal{W}$ that maps (real and artificial) samples to a real number that normally corresponds to some estimate of the distance between $\Prob_\text{g}$ and $\Prob_\text{r}$, for example, the Kullback-Leibler divergence or Wasserstein distance.
Then we set up a two-player game, wherein we switch off between $f_w$ learning the distance between $\Prob_\text{g}$ and $\Prob_\text{r}$ and $g_\theta$ taking an (unbiased) gradient step to decrease that distance.

\subsection{WGAN}
We now describe the Wasserstein GAN (WGAN) formulation of GAN training~\citep{arjovsky2017wasserstein}, beginning with the definition of the earth-mover distance between two probability distributions.
\begin{definition}[Earth-Mover (EM) Distance]
The Earth-Mover (EM) distance or Wasserstein-1 is defined as
\[
W(\Prob_\text{r},\Prob_\text{g}) \coloneqq \inf_{\gamma\in\Pi(\Prob_\text{r},\Prob_\text{g})} \Expect_{(x,y)\sim\gamma}\left[\|x-y\|\right]
\]
where $\Pi(\Prob_\text{r},\Prob_\text{g})$ denotes the set of all joint distributions $\gamma(x,y)$ whose marginals are $\Prob_\text{r}$ and $\Prob_\text{g}$.
We will refer to this quantity as the EM distance and Wasserstein distance interchangeably.
\end{definition}
Roughly, the Wasserstein distance corresponds to the amount of ``effort'' required to transform one probability distribution to the other.
One major benefit of the EM distance is that it is well-defined when the support of the distributions are non-overlapping.
By the Kantorovich-Rubinstein duality~\citep{villani2008optimal}, we know that the Wasserstein distance 
\begin{equation}
W(\Prob_\text{r},\Prob_\text{g}) = \sup_{\|f\|_L \leq 1} \Expect_{x\sim\Prob_\text{r}} [f(x)] - \Expect_{x\sim\Prob_\text{g}} [f(x)]
\label{eq:wasserstein}
\end{equation}
where the supremum is over all the $1$-Lipschitz functions $f:\mathcal{X}\rightarrow\reals$.
Therefore, the optimal value of the following optimization problem
\begin{equation}
\begin{aligned}
& \underset{w \in \mathcal{W}}{\text{maximize}}
& & \Expect_{x\sim\Prob_\text{r}} [f_w(x)] - \Expect_{x\sim\Prob_\text{g}} [f_w(x)]\\
& \text{subject to}
& & \|\nabla_x f_w(x)\|_2 \leq 1\; \text{for all} \; x \in \mathcal{X}
\end{aligned}
\label{eq:problem1}
\end{equation}
is the Wasserstein distance, provided the optimal value satisfies the $\sup$ in \eqref{eq:wasserstein}.
Here, we have used the fact that a differentiable function is 1-Lipschitz if and only if it has a gradient norm of at most 1 everywhere.
If we solve \eqref{eq:problem1}, we can then take stochastic gradients of the Wasserstein distance, because
\begin{equation}
\nabla_\theta W(\Prob_\text{r},\Prob_\text{g}) = - \Expect_{z \sim p(z)} [\nabla_\theta f(g_\theta(z))].
\label{eq:grad}
\end{equation}
See Theorem 3 in~\citep{arjovsky2017wasserstein} for a proof.
The main issue with this process is that the gradient constraint in \eqref{eq:problem1} is challenging to enforce, because it needs to be satisfied for all $x$ in $\mathcal{X}$, but a compromise is to use what is called the \emph{gradient penalty} method~\citep{gulrajani2017improved}.
Our problem then becomes the unconstrained problem
\begin{equation}
\begin{aligned}
& \underset{w \in \mathcal{W}}{\text{maximize}}
& & \Expect_{x\sim\Prob_\text{r}} [f_w(x)] - \Expect_{x\sim\Prob_\text{g}} [f_w(x)] + \beta \cdot \Expect_{\hat{x}\sim\Prob_{\hat{x}}} [\max(0, \|\nabla_{\hat{x}} f_w(\hat{x})\|_2 - 1)^2].
\end{aligned}
\label{eq:problem2}
\end{equation}
where $\Prob_{\hat{x}}$ is some sampling distribution for $\hat{x}$ and $\beta > 0$ is the penalty parameter.
We can solve this optimization problem with a stochastic gradient method, and then solve the outer optimization problem of minimizing the EM distance by another stochastic gradient method using the gradient in \eqref{eq:grad}.

\subsection{Curriculum WGAN} \label{wgan-c}
In our formulation, instead of fixing one discriminator $f_w$, we consider convex combinations of a pre-defined set of discriminators, with the weighting denoted by $\lambda\in\reals_+^d$ such that $\ones^T\lambda=1$.
This means that our discriminator function can be written as $f_w(x) = \sum_{i=1}^d \lambda_i f_i(x)$. 
We also impose that $f_i\in\mathcal{F}_i$ come from increasingly large function classes, or that $\mathcal{F}_1 \subseteq \dots \subseteq \mathcal{F}_d$.
Intuitively, one can view the weight $\lambda$ as modulating the ``strength'' the discriminator.
One can also interpret $\lambda$ as an \emph{attention} mechanism on the overall discriminator.

We further impose that $\mathcal{F}_i$ is a convex set of functions (note that this is different than a set of convex functions).
\begin{definition}[Convex set of functions]
A set of functions $\mathcal{F}$ is convex if for all $f_1,f_2\in\mathcal{F}$ and $\alpha\in[0,1]$, we have that
\[\alpha f_1 + (1-\alpha)f_2 \in \mathcal{F}.\]
\end{definition}
It will become clear later why we need this assumption, and we will also see that it is satisfied by the neural network discriminators we use.
Given these assumptions, we can now define a partial order on the weight vector $\lambda$.
As we will see, this ordering corresponds to the strength of the discriminator, or equivalently, the difficulty the generator will have in lowering the Wasserstein distance.
\begin{definition}[Partial Ordering on $\lambda$]
We write $\lambda_1\succeq\lambda_2$ if the set of functions $f_w^{\lambda_1} = \{\sum_{i=1}^d\lambda_{1i} f_i \mid f_i \in \mathcal{F}_i\}$ contains the set $f_w^{\lambda_2} = \{\sum_{i=1}^d\lambda_{2i} f_i \mid f_i \in \mathcal{F}_i\}$.
We also equivalently say that $\lambda_1$ dominates $\lambda_2$. If neither $\lambda_1\succeq\lambda_2$ nor $\lambda_1\preceq\lambda_2$, we write $\lambda_1\sim\lambda_2$, meaning that neither $\lambda_1$ nor $\lambda_2$ dominates the other.
\end{definition}
A sufficient condition for $\lambda_1$ to dominate $\lambda_2$ is that $\lambda_1$'s backwards cumulative sum is always greater than $\lambda_2$'s backwards cumulative sum, or
\begin{equation}
\sum_{i=k}^d \lambda_{1i} \geq \sum_{i=k}^d \lambda_{2i}
\label{eq:sufficient}
\end{equation}
for all $k \in \{1, \ldots, d\}$.
The following example illustrates why this is a sufficient condition, at least for $d=2$.
\begin{exmp}
Let
\begin{equation}
\begin{aligned}
f_a(x) &= \alpha f^a_1(x) + (1-\alpha)f^a_2(x)\\
f_b(x) &= \beta f^b_1(x) + (1-\beta)f^b_2(x)
\end{aligned}
\end{equation}
where $f^a_1,f^b_1 \in \mathcal{F}_1$, $f^a_2,f^b_2\in\mathcal{F}_2$, and $\mathcal{F}_1$ and $\mathcal{F}_2$ are convex sets of functions such that $\mathcal{F}_1 \subseteq \mathcal{F}_2$.
Here, $\lambda_a=[\alpha, 1-\alpha]$ and $\lambda_b=[\beta,1-\beta]$ and we have that $\lambda_a \succeq \lambda_b$ if $\alpha \leq \beta$.
Suppose that in fact $\alpha \leq \beta$.
Then let $f^a_1(x)=0$ and $f^a_2(x) = \frac{\beta}{1-\alpha}f^b_1(x) + \frac{1-\beta}{1-\alpha}f^b_2(x)$, which is in $\mathcal{F}_2$ because $\mathcal{F}_2$ contains all convex combinations of functions inside it and $\mathcal{F}_1\subseteq\mathcal{F}_2$.
Then $f_a(x) = (1-\alpha)(\frac{\beta}{1-\alpha}f^b_1(x) + \frac{1-\beta}{1-\alpha}f^b_2(x)) = f_b(x)$ and we have shown that $f_b$ is representable by $f_a$.
\label{exmp:1}
\end{exmp}
Continuing the logic in Example~\ref{exmp:1} by induction, it is easy to see that~\eqref{eq:sufficient} is a sufficient condition.

\begin{algorithm}[t!]
  \caption{Curriculum WGAN}
  \label{alg:1}
  \begin{algorithmic}[1]
    \Require $\alpha$, optimization algorithm parameters. $\beta$ gradient penalty parameter. $m$, the batch size. $n_\text{critic}$, number of inner critic iterations.
    \Require $w_0$, $\theta_0$, initial network parameters. $\pmb{\lambda} = [\lambda^{(1)}, \lambda^{(2)}, \dots]$, a sequence where $\lambda^{(i)}\preceq\lambda^{(i+1)}$.
    \While{$\theta$ has not converged}
    \State $\lambda \leftarrow$ next$(\pmb{\lambda})$.
    \State Let $f_w(x) = \sum_{i=1}^m \lambda_i f_i(x)$.
    \For{$t=0,\ldots,n_\text{critic}$}
    	\State Sample $\{x^{(i)}\}_{i=1}^m\sim\Prob_\text{r}$ a batch of real data.
        \State Sample $\{z^{(i)}\}_{i=1}^m\sim p(z)$ a batch of prior samples.
        \State Sample $\{\epsilon^{(i)}\}_{i=1}^m\sim U[0,1]$ a batch of random weights
        \State $\tilde{x}^{(i)} \leftarrow g_\theta(z^{(i)})$
        \State $\hat{x}^{(i)} \leftarrow \epsilon x^{(i)} + (1-\epsilon) \tilde{x}^{(i)}$
        \State $g_w \leftarrow \nabla_w [\frac{1}{m} \sum_{i=1}^m f_w(\tilde{x}^{(i)}) - f_w(x^{(i)}) + \beta \max(0, \|\nabla_{\tilde{x}} f_w(\hat{x}^{(i)})\|_2 - 1)^2]$
        \State $w \leftarrow \text{update}(w, g_w, \alpha)$
    \EndFor
    \State Sample $\{z^{(i)}\}_{i=1}^m\sim p(z)$ a batch of prior samples.
    \State $g_\theta \leftarrow \nabla_\theta [\frac{1}{m} \sum_{i=1}^m -f_w(g_\theta(z^{(i)}))]$
    \State $\theta \leftarrow \text{update}(\theta, g_\theta, \alpha)$
    \EndWhile
\end{algorithmic}
\end{algorithm}

The WGAN Curriculum (WGAN-C) algorithm that we propose is summarized in Algorithm~\ref{alg:1}. It is essentially the Improved WGAN algorithm~\citep{gulrajani2017improved}, altered to include the curriculum as defined by an increase of $\lambda$ on each iteration to strengthen the discriminator over the course of training (although we include the Wasserstein GAN algorithm here, this technique is also compatible with the original GAN objective).
The basic idea is to define a curriculum of increasingly difficult $\lambda$s, made quantitative by the constraint that $\lambda_i \preceq \lambda_{i+1}$.
Our thesis is that slowly unshackling the discriminator, and thus increasing the difficulty of the learning task presented to the generator, will lead to a more stable learning algorithm.
(See the Supplementary Materials for a rough connection of our method to trust region methods in optimization.)

We are now able to rigorously define the strength of the discriminator, or equivalently the difficulty level for the generator.

\begin{definition} [$\varepsilon$-fooling the discriminator]
We say that a generator $\varepsilon$-fools a discriminator $f_w$ for some $\varepsilon>0$ if the optimal value of the maximization problem in \eqref{eq:problem1} is less than $\varepsilon$. That is to say, if the learned discriminator $f_w$ results in a Wasserstein distance of less than $\varepsilon$, we consider it $\varepsilon$-fooled.
\end{definition}

\begin{theorem} [Generator Curriculum]
Suppose $\lambda_i \preceq \lambda_{i+1}$ and $\mathcal{F}_1 \subseteq \dots \subseteq \mathcal{F}_d$.
If $f_w^{\lambda_{i+1}}$ is $\varepsilon$-fooled by a generator $G$, then $f_w^{\lambda_i}$ is also $\varepsilon$-fooled by $G$.
\label{thm1}
\end{theorem}
Proof: Since the set $\{\sum_{j=1}^d\lambda_{i,j} f_i \mid f_j \in \mathcal{F}_j\} \subseteq \{\sum_{j=1}^d\lambda_{i+1,j} f_j \mid f_j \in \mathcal{F}_j\}$ the maximum value attained when optimizing over the latter set will necessarily be greater than the former. Thus, if the optimal discriminator in the latter set is $\varepsilon$-fooled, then the optimal discriminator in the former set is necessarily $\varepsilon$-fooled.

This means that at iteration $i$ of the algorithm, the discriminator can produce a lower bound of the Wasserstein distance, which we denote $p^i$.
Because of our ordering of discriminators, we have that $p^i$ are monotonically increasing throughout the algorithm, or $p^i \leq p^{i+1}$ by Theorem~\ref{thm1} for a fixed generator.
Thus, even though at iteration $i$ we calculate an (approximate) Wasserstein distance, our generator can safely minimize $p^i$ at each iteration, because it is necessarily a lower bound on the actual Wasserstein distance.
Because we are always minimizing a lower bound on the true objective, this makes the optimization algorithm more stable.
Intuitively, as more capacity is allowed to the discriminator throughout the algorithm, the task of minimizing the lower bounds becomes more and more challenging.
However, the generator has already (approximately) minimized all of the previous lower bounds, so it is well suited for the harder task.

\section{Experiments} \label{experiments}
We evaluate the Curriculum WGAN on a sinusoid generation task and a celebrity image synthesis task.
We refer the reader to Section~\ref{sec:designing-discriminators} in the Supplementary Materials for more details on how to design a curriculum of generators for images and time series data, as well as an argument for why neural networks form a convex set of functions.

\subsection{Sinusoids} \label{sinusoids}
In the sinusoid generation task, the goal is to generate one-dimensional sine waves, that is, $x_t = A\sin(\omega t + b)$.
In this case, the generator and discriminators are two layer neural networks with $128$ hidden units.
The generator is attempting to output length $64$ sine waves.
Assuming the original length of the time series is $T$, we define a sequence of discriminators acting on successively longer parts of the input for $i=1,\ldots,T$.
This guarantees that $\mathcal{F}_i \subseteq \mathcal{F}_{i+1}$, as the $i$th discriminator can only act on fewer points of $x$.
There are lots of options, then, for the curriculum of $\lambda$s.
The simplest schedule (that we use in our experiments) is to set $\lambda=e_1$ for some fixed number of iterations, then set $\lambda=e_2$ for the same number of iterations, all the way up to $\lambda=e_T$.
This also has the advantage of sparsity; at any one time we are updating only one discriminator.
In our experiments, when we update $\lambda$, we simply randomly initialize a new discriminator network with the appropriately sized input.

\begin{figure}
\vspace{-20pt}
\includegraphics[scale=0.2]{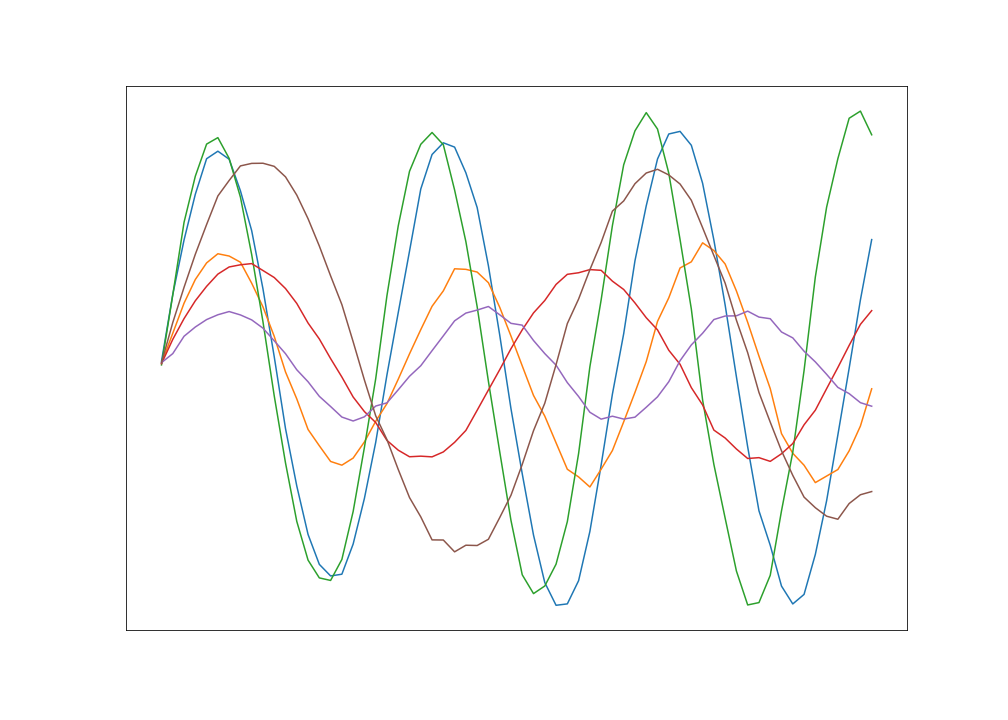}
\includegraphics[scale=0.2]{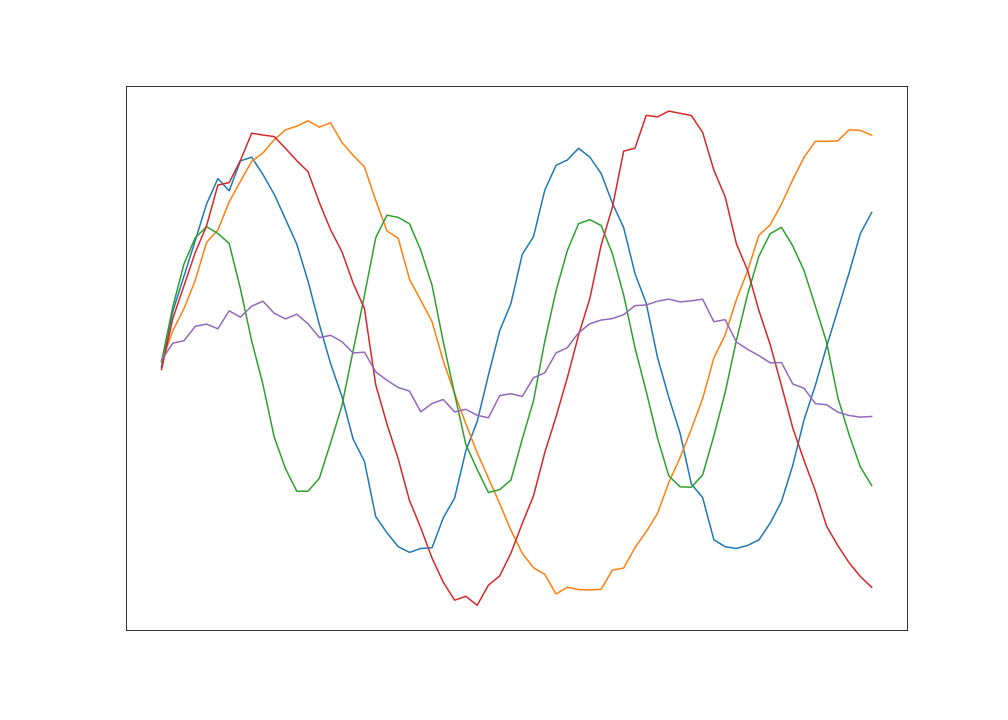}
\vspace{-20pt}
\caption{The plot on the left shows sinusoids generated with a progressive lengthening strategy to create a curriculum GAN. On the right are sinusoids generated with the same setup but no progressive lengthening. The generated sinusoids improve significantly when discriminator attention is grown (via progressive lengthening) instead of training the discriminator on full length sequences to start.}
\label{fig:gen-sin}
\end{figure}
We run the WGAN-C algorithm with the $\lambda$ schedule as described above, and contrast it to the WGAN algorithm (without curriculum).
In fact, WGAN-C runs much faster, as the sequence length increases throughout training.
Output from WGAN-C is shown side-by-side with WGAN in Figure~\ref{fig:gen-sin}. It is easy to see visually that WGAN-C out performs WGAN.
To the best of our knowledge, no experiment on progressive lengthening of time-series data has been undertaken to date.
We note that a similar experiment was performed in~\citep{esteban2017medical}, but without a curriculum.
We have also run the same experiments with the original GAN objective~\citep{goodfellow2014gans}, and the results indicate that the improvements afforded by curriculum training are present independent of the choice of GAN training algorithm.

In addition to the visual comparison, we measured the average $\ell$-2 error of the generated waves to the closest sine wave in the dataset (by discretizing the range of sinusoids that generate the dataset). At the end of training, the average error of sinusoids generated by a progressive lengthening strategy is 33.6\% lower---the average minimum $\ell$-2 distance from an element the training dataset is $1.13\pm0.01$ for the sinusoids generated by a growing strategy, and the error is $1.51\pm0.06$ for those generated without a growing strategy.

\subsection{CelebA-HQ}\label{celeba}

\begin{figure}
\vspace{-10pt}
\centering
\includegraphics[scale=0.35]{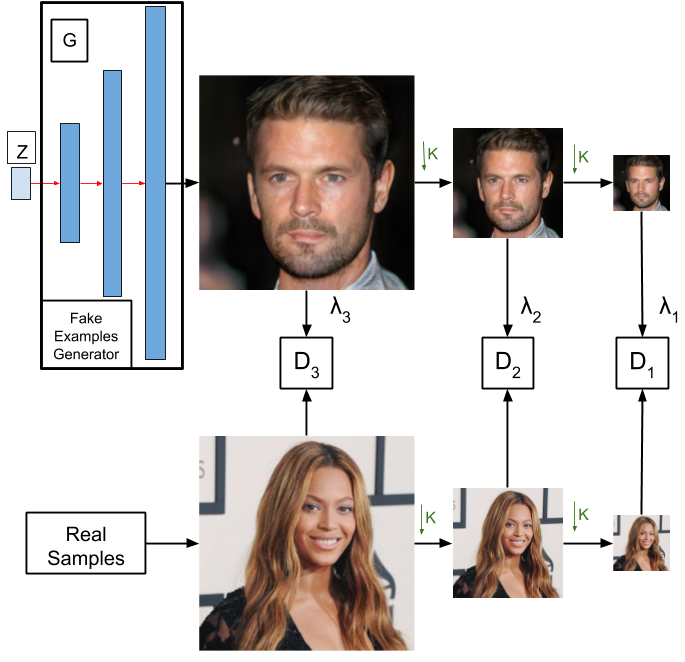}
\caption{Our training setup for Celeba-HQ. $D_2$ and $D_1$ operate on successively downsampled versions of the real and fake images. In this figure, each downsample operation reduces the image size by a factor of $K$. Our curriculum uses 5 discriminators, and the downsampling factor $K$ is 2. Thus, the discriminators range from operating on 64 $\times$ 64 to 4 $\times$ 4 images.}
\label{fig:image-setup}
\end{figure}

The image synthesis task provides a direct comparison to the work of~\citet{karras2017progressive}, which uses a progressive growing strategy to deliver state-of-the-art image synthesis results using the Celeba-HQ dataset.
We duplicate their training setup on CelebA-HQ, replacing their strategy of progressively growing the size of the generator and discriminator network with our simpler method of progressively growing the discriminator attention to create a curriculum. We achieve very similar results, which indicates that their method is a special case of our more general technique. See discussion in Section~\ref{discussion-related} for more.

The CelebA-HQ dataset is a refinement of the CelebA dataset \citep{liu2015faceattributes} that consists of high-quality celebrity images that have been centered and cropped to maximize the view of the face. 
We use the training setup depicted in Figure~\ref{fig:image-setup} in which the curriculum is determined by the image average downsampling operator defined in~\ref{downsampling}. We have a separate discriminator for each successively downsampled version of the final image, including at $4 \times 4, 8 \times 8, 16 \times 16, 32 \times 32,$ and $64 \times 64$. 
The full image size output by the generator is $64 \times 64$. We begin with $\lambda = [1, 0, 0, 0, 0]$, meaning that the effective discriminator only considers $4 \times 4$ downsampled versions of the generated outputs and training data. We slowly change $\lambda$ to focus on higher fidelity versions of the images.

Throughout training, we follow an identical schedule in switching to discriminators that operate on larger images (less downsampled versions of the output) as \citep{karras2017progressive} use to progressively grow the output size of their GAN, and our results come out substantially similar. Our findings suggest that their results are achieved by the effective use of a curriculum to slowly build the strength of the generator, rather than from the more complicated layerwise pretraining aspect of their method.

\begin{figure}
\vspace{-20pt}
\centering
\includegraphics[scale=0.4]{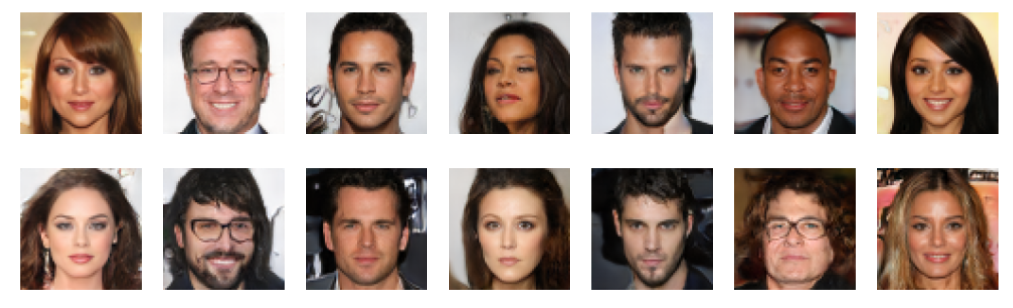}
\caption{The bottom row shows the final output of \citep{karras2017progressive} for 64$\times$64 images, and the top row shows our output images. These images were not cherry picked for quality, though some especially bad images were discarded when randomly selecting.}
\label{fig:celebaHQ}
\end{figure}

\begin{figure}
\vspace{-20pt}
\centering
\includegraphics[scale=0.4]{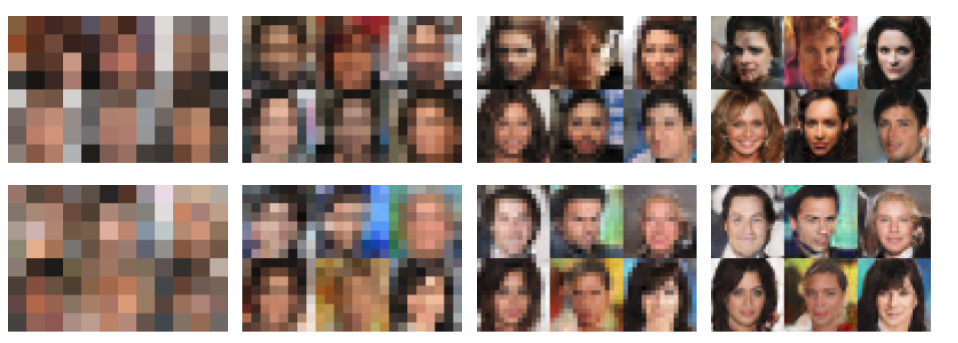}
\caption{The bottom row shows the converged smaller images generated by \citep{karras2017progressive} over the course of training. The top row shows our corresponding downsampled images at the same points in training. Our methods perform very similarly throughout the training process, in addition to having similar outputs.}
\label{fig:celebaHQ-training}
\end{figure}

The bottom row in Figure~\ref{fig:celebaHQ} shows the results of \citep{karras2017progressive}, and the top row shows our results for 64$\times$64 images. In Figure~\ref{fig:celebaHQ-training}, the bottom row is the converged smaller images generated by \citep{karras2017progressive}, and the top row is our corresponding downsampled images.

We would note that~\citet{karras2017progressive} were able to output $1024 \times 1024$ images, due in part to the sheer quantity of images their network was able to see, as a consequence of their method beginning with smaller-sized outputs and gradually increasing output size. They are able to train with much larger batch sizes in the early stages when the output size of the network is small. We output $64 \times 64$  images in order to obtain our results in a reasonable amount of time (4 days of training on an NVIDIA GeForce GTX 1080 Ti). Due to the batch size constraints imposed by GPU memory, our method, which always outputs full-sized images, takes comparatively longer to produce final outputs. In regimes where GPU memory is not highly restrictive, our method should perform equally well at generating larger sized images on a similar time scale. We have initiated $256 \times 256$ image experiments and plan to include the results as soon as they are available; for comparison purposes, the \citep{karras2017progressive} experiments took 20 days to run on a NVIDIA Tesla P100.

\section{Discussion and Related Work} \label{discussion-related}
\citet{bengio2009curriculum} introduced curriculum learning in the context of machine learning, formalizing the intuition that agents learn better when presented with a curriculum, i.e. a series of tasks of increasing difficulty. Their work primarily explored curriculum strategies in the context of classification and sequence prediction, showing that curriculum strategies help more quickly find local minima of non-convex loss functions. Curriculum strategies have since found particularly widespread use in training recurrent neural networks \citep{zaremba2014execute,bengio2015schedule}, and it is typical to see some variant of ``teacher forcing'' or ``scheduled sampling'' in applications where recurrent architectures are used. 

Although it was not explicitly expressed as such, we view \citep{karras2017progressive} in part as an application of curriculum learning to the training of a GAN in the context of image modeling. This view motivated our work to strip extraneous elements from the progressive growing strategy and test our framework.

When GANs have been applied to language modeling, they have frequently inherited the recurrent neural network architectures common in natural language tasks. As such, \citet{press17language-gen} use a recurrent architecture for language generation, which combines the use of a GAN with a progressive lengthening curriculum for the task of language generation. We see this as another instantiation of our framework of GAN curricula, contributing to the success of their model on the language generation task. \citet{fedus18maskgan} also uses a recurrent architecture along with a GAN for the task of filling in the blank in sentences. The size of the blank in the sentence is grown over the course of training, and ultimately the language model simply outputs natural language. This model is made slightly more complex with the use of reinforcement learning for training on a discrete output space \citep{sutton1998reinforcement-learning}, which makes it less easy to understand as a pure instantiation of our framework, but the method nonetheless makes significant use of a curriculum strategy.

Though these works in natural language generation used a variant of curriculum learning during training, the method has largely been inherited along with the recurrent networks they use, despite not being properly motivated in the context of GANs. The training setup of GANs bears little resemblance to the supervised curriculum learning studied in \citep{bengio2009curriculum}, which further motivated our work to establish empirically the efficacy of a curriculum strategy when using GANs.

Our work may also be connected to state-of-the-art work in text-to-image-synthesis, though the link is more tenuous. \citet{zhang2017stackgan++} uses multiple discriminators \emph{and} multiple generators for different components of the text-to-image task, breaking down the generation process into easier constituent parts and tasking a separate GAN to learn each component. The reason this method works may be connected to the reason curriculum strategies are effective: that they start by teaching the generator simpler tasks and build upon the initial successes to formulate the final outputs.

As discussed earlier,~\citet{lucic2017gans} demonstrated in their recent paper that GAN improvements in areas such as image generation cannot be universally applied to improve GANs for other purposes. Though the value of improving GANs for specific domains should not be diminished, it would obviously be more desirable to find general principles for improving all GANs. With an eye to discovering these general principles, our results provide a clear demonstration of the positive effects of curriculum learning on image generation by GANs. Along with the prevalence of curriculum learning in training methods that have shown success across other data modalities, our formalization of curriculum learning as a standalone training method for GANs calls for further investigation into its relevance to GANs generally. We intend to develop training procedures based on curriculum GANs in other contexts, with the aim of uncovering a broadly applicable training method. Thus, we hope that we have taken an important first step towards meeting the challenge that~\citet{lucic2017gans} present: to discover training methods that can be applied to enhance all GANs.

% This material is based upon work supported by the National Science Foundation under Grant No. 2017245257.

\small

\bibliography{nips_2018}
\bibliographystyle{apalike} %unsrt?

\newpage
\begin{center}
{\bf \huge Supplementary Materials}
\end{center}

\begin{appendix}
\section{Designing Discriminators}
\label{sec:designing-discriminators}

WGAN-C requires sets of functions $\mathcal{F}_1,\ldots,\mathcal{F}_d$ such that $\mathcal{F}_i$ is a convex set of functions and $\mathcal{F}_i\subseteq\mathcal{F}_{i+1}$.
Normally, these sets will be neural networks operating on different versions of the input, however, any variation that satisfies the constraints we lay out fits in our framework.

We now show that when $f_i$ is a neural network, $\mathcal{F}_i$ is a convex set of functions, assuming our neural networks have enough capacity.
Recall the universal approximation theorem, \ie,  a neural network with enough capacity can approximate any continuous (and hence differentiable) function arbitrarily well on a closed, bounded subset of $\reals^n$~\citep{hornik1989multilayer}.
Because $\mathcal{F}_i$ includes all continuous functions, and because the sum of a finite number of continuous functions is a continuous, we can conclude that $\mathcal{F}_i$ is a convex set of functions.

We now describe how our framework can (and has already been in disguise) applied to several application areas.

\subsection{Images}
When $x$ is an image, we have that $x\in[0,1]^{W\times H\times C}$ where $W\in\mathbb{N}_+$ is the width, $H\in\mathbb{N}_+$ is the height, and $C\in\mathbb{N}_+$ is the number of channels.
One option for our discriminators, in this case, is for them to be (deep) convolutional networks applied to \emph{downsampled} versions of the image.
\begin{definition}[Image Average Downsampling] \label{downsampling}
Given an image $x\in[0,1]^{W\times H\times C}$, its downsampled version $D_k(x)$, defined for $W \mod k = 0$ and $H \mod k = 0$, is given by taking the average of every $k \times k$ non-overlapping block of the image. Thus, we have that $D_k(x) \in [0,1]^{W/k \times H/k \times C}$.
\end{definition}

Assuming that our image dimensions are the same ($W=H$) and we have that $W \mod 2=0$, which can be accomplished with interpolation, we can define a sequence of downsampled images, given by $D_{W}(x), D_{W/2}(x), \ldots, D_2(x), D_{1}(x)$.
The first of these $D_{W}(x)$ is the average of the entire image, and the last of these $D_{1}(x)$ is the original image.
Our discriminators, then, can be (deep) convolutional networks applied to each of these downsampled images.
It is important to note that image average downsampling is differentiable, which means that we can backpropagate through it.
An illustration of this training pipeline is displayed in Figure~\ref{fig:image-setup}.

\subsection{Sequences}
In the domain of natural language, a natural way to design a curriculum for the generator is to train it to produce progressively longer sequences of words that are processed by the discriminator.
It is already common to use this strategy when the generator is implemented by a recurrent architecture, but this holds independent of the architecture of the generator.
With time series data, a natural way to design a curriculum for the generator is to train it to produce progressively longer time series that are processed by the discriminator.

\section{Connection to Trust-Region Methods}
Trust-region methods in numerical optimization define a bounded region around the current iterate within which they trust a simplified model to be an adequate representation of the objective function, and then choose an update that is the approximate minimizer of the model in this trust region~\citep{nocedal2006}.
Our method is (roughly) a trust-region method for GAN training, as we gradually increase the class of functions (the trust region) our discriminator can take on throughout training, which in turn increases the possible gradient updates for the generator.

\section{Sinusoid Generation Wasserstein Distance Plots}

\begin{figure}
\centering
\includegraphics[width=.4\columnwidth]{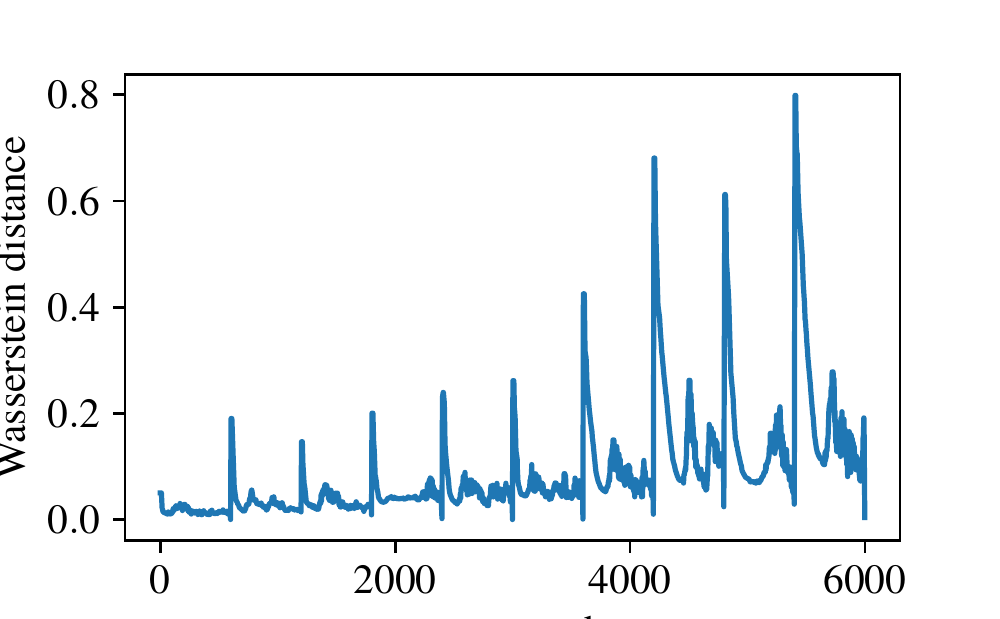}
\caption{Wasserstein Distance over the course of training for WGAN-C. Note the spikes once $\lambda$ is changed.}
\end{figure}

\begin{figure}
\centering
\includegraphics[width=.4\columnwidth]{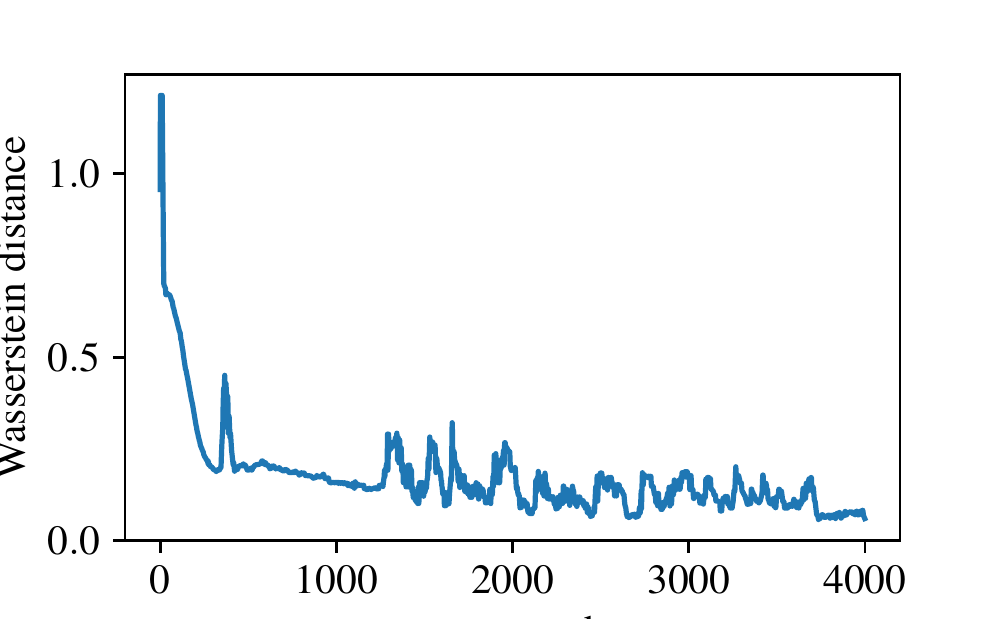}
\caption{Wasserstein Distance over the course of training for standard WGAN.}
\end{figure}

\end{appendix}

\end{document}